\documentclass[conference]{IEEEtran}
\usepackage{cite}

\ifCLASSINFOpdf
   \usepackage[pdftex]{graphicx}
\else
\fi
\usepackage[cmex10]{amsmath}
\usepackage{amsfonts} 
\usepackage{array}

\usepackage{mdwmath}
\usepackage{mdwtab}
\usepackage{url}

\renewcommand{\figurename}{Figure}

\begin{document}
%
\title{Online Learning for Ground Trajectory Prediction}

%
\author{
\IEEEauthorblockN{Areski Hadjaz}
\IEEEauthorblockA{Thales Air Systems, Rungis, France\\
Email: areski.hadjaz@thalesgroup.com}
\and
\IEEEauthorblockN{Ga\'etan Marceau}
\IEEEauthorblockA{Thales Air Systems and TAO, Rungis, France\\
Email: gaetan.marceaucaron@thalesgroup.com}
\and
\IEEEauthorblockN{ \ \ \ \ \ \ \ \ \ \ \ \ \ \ Pierre Sav{\'e}ant}
\IEEEauthorblockA{ \ \ \ \ \ \ \ \ \ \ \ \ \ \ \ \ Thales Research \& Technology, Palaiseau, France\\
\ \ \ \ \ \ \ \ \ \ \ \ \ \ \ \ \ Email: pierre.saveant@thalesgroup.com}
\and
\IEEEauthorblockN{ \ \ \ \ \ \ \ \ \ \ \ \ \ \ Marc Schoenauer}
\IEEEauthorblockA{ \ \ \ \ \ \ \ \ \ \ \ \ \ \ TAO, INRIA Saclay, France\\
 \ \ \ \ \ \ \ \ \ \ \ \ \ \ Email: marc.schoenauer@inria.fr}
}




\maketitle

\begin{abstract}
This paper presents a model based on an hybrid system to numerically simulate the climbing phase of an aircraft. 
This model is then used within a trajectory prediction tool. 
Finally, the Covariance Matrix Adaptation Evolution Strategy (CMA-ES) optimization algorithm is used to tune five selected parameters, and thus improve the accuracy of the model.
Incorporated within a trajectory prediction tool, this model can be used to derive the order of magnitude of the prediction error over time, and thus the domain of validity of the trajectory prediction.
A first validation experiment of the proposed model is based on the errors along time for a one-time trajectory prediction at the take off of the flight with respect to the default values of the theoretical BADA model. 
This experiment, assuming complete information, also shows the limit of the model.
A second experiment part presents an on-line trajectory prediction, in which the prediction is continuously updated based on the current aircraft position. 
This approach raises several issues, for which improvements of the basic model are proposed, and the resulting trajectory prediction tool shows statistically significantly more accurate results than those of the default model. 
\end{abstract}




%
\IEEEpeerreviewmaketitle

\section{Introduction}
\label{sec:intro}
Trajectory Prediction (TP) is the core component of automated systems in Air Traffic Control (ATC). 
Many functionalities of the Decision Support Tool directly rely on an accurate TP: controller posting, workload estimation, arrival sequencing, loss of separation detection, and conflict resolution, to name the most prominent ones. 
As a consequence, TP is the weakness of the current automation ATC systems and a major issue in the ATC research community, even more with the new paradigmatic shift toward 4D trajectories in both the SESAR Joint Undertaking and the NextGen project.
The challenge is to reduce the uncertainty of the prediction of the aircraft states on a temporal horizon of at least 20 minutes. 
To achieve this, the information of the current state of the aircraft and its environment shall be reliable.
As a matter of fact, the Flight Management System (FMS) has access to the measurements from the sensors of the aircraft and creates its own TP, which is updated frequently. 
Therefore, we should expect that this TP is the most accurate one.
Thanks to the Data Link, it should be possible for the ground control to receive data from this on-board TP. 
Unfortunately, today, this promising technology is not implemented in all aircrafts. 
Moreover, a decision support tool requires the capacity to efficiently generate 'what-if' scenarios. 
Today, it is unrealistic to receive many trajectories from the on-board system of every aircraft in the airspace and merge them into different airspace scenarios in real-time. 
Therefore, the ground TP is still an essential component of the future air traffic control systems.

A trajectory can roughly be separated in three phases: climbing, level flight, and descent. 
Phases with altitude changes are the most difficult to handle due to the differences of aircraft performances and the effect of the weather conditions. 
Operationally, the controllers of approach control centers, who deal with climbing and descent phases, must have a good representation of the vertical evolution of the aircraft to ensure their separations. 
In most airspace, this separation shall be at least 10 flight levels. 
For these reasons, this study will focus mainly on the prediction in the vertical axis.

The paper is organized as follows: Section \ref{sec:relatedWork} surveys some recent works in the domain of trajectory prediction. 
Section \ref{Model} details the proposed model. 
The validation experiments are described in Section \ref{sec:modelValidation}, while the on-line prediction method is experimented with in Section \ref{sec:Online}. 
Section \ref{conclusion} summarizes the paper and gives some hints for further research in the area of trajectory prediction.

\section{Related Work}
\label{sec:relatedWork}
Research on TP includes a set of methodologies for its specification, implementation and evaluation.
First, the Eurocontrol Specification for Trajectory Prediction gives the requirements for an operational TP and a validation methodology useful for the Air Navigation Service Providers in their choice of a new ATM system compatible with SESAR. 
This document covers many functionalities like flight plan and clearance processing, airspace constraints and real-time monitoring. 
Unfortunately, it does not specify the accuracy requirements. 
Besides, \cite{Musialek2010} provides an important literature survey of trajectory prediction technology with 282 reviewed documents and 20 selected for further studies. 
From the selected set, many implement the point-mass model where the rotational moments are not modeled. 
This is an acceptable assumption for airline aircraft. 
Also, \cite{Coppenbarger1999} and \cite{Swierstra2003} enumerate the principal difficulties inherent to TP, i.e. the uncertainty on the input data, the controller and pilot intents, and quantify the errors accordingly. 
The input data mainly refers to aircraft characteristics which are given in the {\it Base of Aircraft DAta} (BADA) Aircraft Performance Model. 
Still, these values are only nominal and can be different from the real situation. 
Therefore, the SESAR project 5.5.2 \cite{SesarTP2012} concludes that sharing airline operational control data, like the mass and the speed schedule, could provide quick improvement to ground trajectory prediction, with limited investment. 
As a matter of fact, these parameters are determinants for computing the TP with a point-mass model.

More generally, a parametric approach refers to a model based on flight equations and aircraft characteristics. 
The point-mass model is an example of a parametric approach. 
As we will see, many parameters can be used to tune the model to reality. 
To generate trajectories, \cite{Gallo} uses six ordinary differential integrators depending on the longitudinal motion instruction. 
Similarly, \cite{Glover2004} and \cite{Kamgarpour2011} use hybrid systems to model the change of differential equations according to the control law and the aircraft states. 
A discrete space, e.g. an automaton, is defined to represent the modes of the system. 
Every mode defines the differential equations and creates trajectories in a continuous space. 
As seen in \cite{Glover2004}, this model is well-suited for implementing BADA. 
\cite{Alligier} exposes a technique to find a general thrust settings, i.e. a control law, that could be used in such framework. 
The idea of fitting the mass parameter of BADA on a few past points is used. 
Results on the accuracy of this generic control law in the vertical plane should be expected in a near future.

\begin{figure}[!h]
\centering
\includegraphics[width=2.5in]{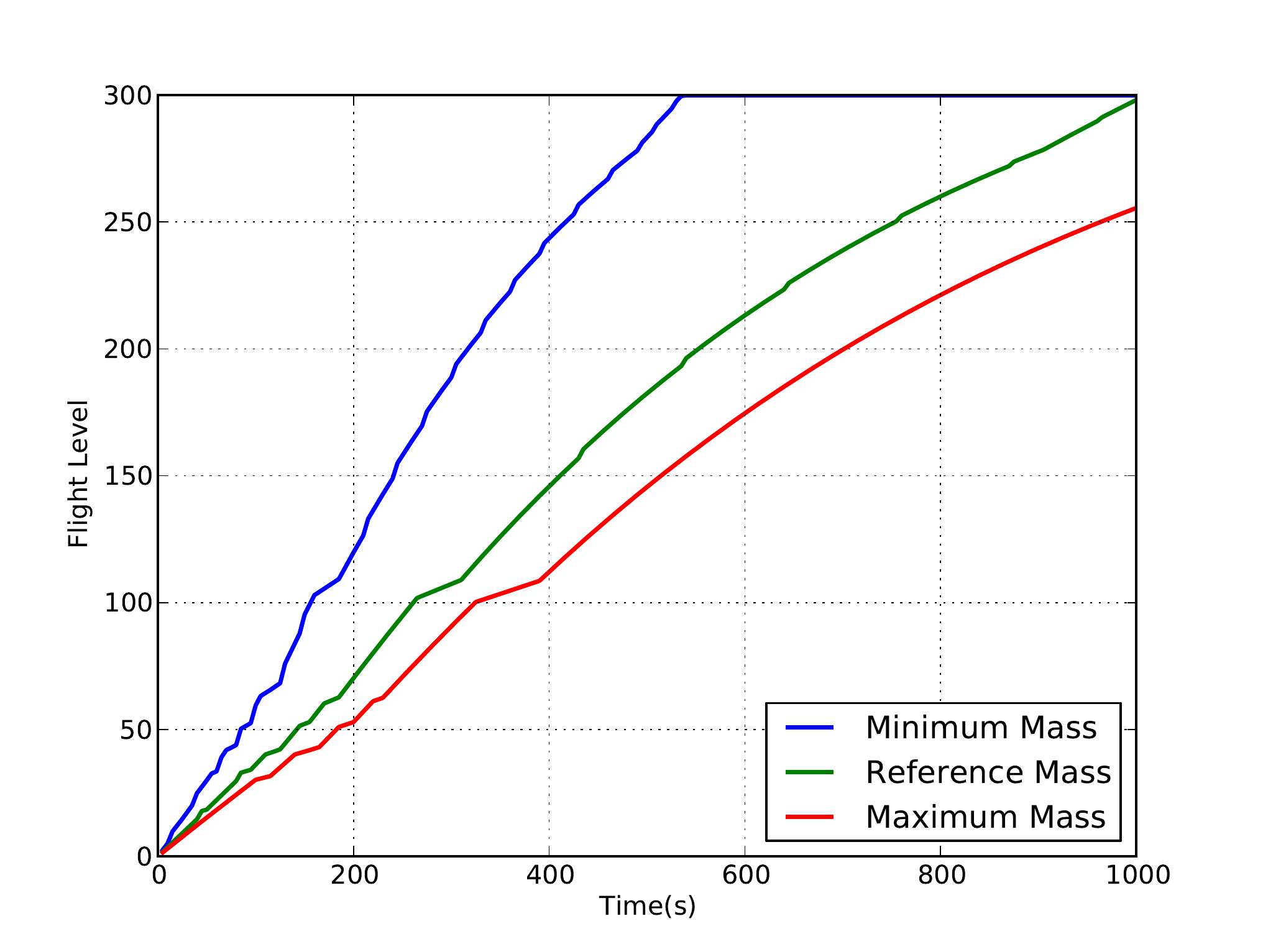}
\caption{The Effect of the mass parameter}
\label{fig:massEffect}
\end{figure}

A common flaw in parametric approaches concerns the nominal values used for every parameters. As shown on figure \ref{fig:massEffect}, the effect of the mass parameter clearly shows the importance of tuning these parameters. 
As an example, 400 seconds after the takeoff, there is already a difference of approximately 250 flight levels between the minimal and the maximum mass which is enormous for an application like the TP. 
This can easily become a burden when the model is rather complex. 
To overcome this inherent difficulty, non-parametric approaches are studied in order to obtain an aircraft model from the past trajectories. 
Non-parametric approaches rely on machine learning and statistical inference: \cite{LeFablec1999} uses neural networks, \cite{M2R_Richard_ALLIGIER} uses genetic programming in order to learn the structure of the variables of a linear regression and \cite{M2R_GHASEMI} uses fuzzy regression with k-nearest neighbor. 
The main drawback of non-parametric approaches is that it requires lots of historical data and the model is learned for a specific context and can hardly be generalized because of the airspace constraints, e.g. aircraft following a Standard Instrument Departure.

Combining both approaches is an interesting research question addressed by \cite{Alligier} and \cite{Crisostomi2008}. 
The latter combines Monte-Carlo Simulation and worst-case scenario for modifying the parameters of BADA while integrating a wind model. 
However, this work is limited to the descent phase and the experiments are performed on trajectories obtained by simulation.
The main contribution of our work on this question is to show that tuning the parameters of BADA during the progress of a flight is a way to overcome the difficulties of parametric approach and avoiding to learn a model from the historical data.
To the best of our knowledge, no work has answered to this question with results of the vertical plane.



\section{Model}
\label{Model}
The basic idea is to create an hybrid system to generate a trajectory in the vertical plane using BADA. The system describes the transitions between the different modes in function of the states and the control laws. Following the mode, a given differential equation is integrated to obtain the trajectory. This section states the assumptions, defines the model and gives an integration schema for generating the trajectories.

\subsection{Assumptions}
Depending on the requirements of the simulation, one may choose different levels of complexity for the coordinate systems and the aircraft control system. Since we consider only the climbing phase, the system will evolve on a short period of time on a limited geographical area. From these considerations, we assume that the flat earth model is a reasonable approximation (cf. \cite{Hull2007}). This model assumes that the earth is flat, non-rotating and can be defined as an inertial reference frame. Also, the gravitational force is constant and perpendicular to the ground. The atmosphere is at rest relative to the earth and atmospheric properties solely depend on the altitude. With these assumptions, one can define the differential equations of kinematics and dynamics. For the kinematics, the equations imply solely that the displacement is proportional to the speed projected on the vertical or horizontal plan. For the dynamics, we need to model the thrust, the drag, the lift and the weight. To our knowledge, BADA is the most complete model describing the capabilities of many aircraft types. 

\subsection{Total-Energy Model}
This section relies on the BADA 3.10 User Manual \cite{Bada2012}. BADA is based on the total-energy model where the rate of work is equal to the rate of potential and kinetic energy. Here, the rate of climb is obtained by controlling the speed and the throttle. From \cite{Bada2012} p.14:eq.3.2-7 and by transforming the quantities into functions of altitude $h$, speed $V$ and mode $q$, the rate of climb is defined by: 
\begin{equation}
  \label{eq:roc}
  \dot h(V,q) =  \frac{T(h) - \Delta T}{T(h)} \left[ \frac{(Thr(h) - D(V,h)) \cdot V}{m \cdot g} \right] \cdot f(V,h,q)
\end{equation}  
where the function $T(h)$ is defined in BADA 3.10 User Manual \cite{Bada2012} p. 10, $Thr(h)$ at p.22, $D(V,h)$ at p.20 and $f(V,h,q)$ at p.15-16. As a first approximation, the mass $m$, the gravitational acceleration $g$ and the temperature differential $\Delta T$ are constant during the climb phase. According to \cite{Hull2007}, this is a reasonable assumption since the climb phase lasts around 20 minutes. The terms  $h$, $V$ and $q$ are variables that evolve with the system and one must specify the evolution of the two remaining independent variables. However, the acceleration in function of the aircraft dynamic is not specified in BADA. From \cite{Hull2007}, it is given by: 
\begin{equation}
  \label{eq:acceleration}
  \dot V(h, \dot h) = \frac{1}{m}(Thr(h) - D(V,h) - m \cdot g \cdot \sin(\gamma))
\end{equation}
where $\sin ( \gamma )=\frac{\dot h}{V}$. So, from eq. \ref{eq:acceleration}, we can see that the acceleration evolves independently of the mode given the rate of climb. Also, when $\dot h$ is high, $\sin( \gamma )$ is high and the acceleration $\dot V$ is low. This goes along with the hypothesis the total-energy model. Finally, the next section will present the evolution of the variable $q$.

\subsection{Mode Definition}
In this section, we define the modes and their transitions based on BADA 3.10 User Manual \cite{Bada2012}. First, let $Q = \{CAS,MACH\} \times \{LOW,HIGH\} \times \{DEC,CST,ACC\}$ be the mode space.
The first two modes, transition altitude and tropopause, depend solely on the altitude: 
\[
  q_1(h) = \left\{ 
  \begin{array}{l l}
    \text{CAS} & \quad \text{if $h \le H_{trans}$}\\
    \text{MACH} & \quad \text{otherwise}\\
  \end{array} \right.
\]

\[
  q_2(h) = \left\{ 
  \begin{array}{l l}
    \text{LOW} & \quad \text{if $h \le H_{trop}$}\\
    \text{HIGH} & \quad \text{otherwise}\\
  \end{array} \right.
\]
where $H_{trans}$ is the transition altitude and $H_{trop}$ is the tropopause geopotential pressure altitude. 
Finally, the last feature $q_3$ is the mode of acceleration. The simplest controller of this mode is given by:
\[
  q_3(V,h) = \left\{ 
  \begin{array}{l l}
    \text{ACC} & \quad \text{if $V \le V^*(h) - \epsilon$}\\
    \text{DEC} & \quad \text{if $V \ge V^*(h) + \epsilon$}\\
    \text{CST} & \quad \text{otherwise}
  \end{array} \right.
\]
where ACC is acceleration, DEC is deceleration and CST is constant speed and $V^*$ be a target speed at altitude $h$ where $V$ tends to converge. $V^*$ can be chosen according to the speed schedule defined in the Airline Procedures Model of BADA 3.10 User Manual \cite{Bada2012} p.29, where three target speeds $(V_1,V_2,M)$ are required as parameters. The nominal values can be found in the airline procedure files of BADA. Finally, $\epsilon \in \mathbb{R}$ is a threshold value to avoid jitter. 
Next, the energy share factor function $f$ takes its values according to the following flight conditions: 
\begin{enumerate}
\item Constant $V_M$ above tropopause,
\item Constant $V_M$ below tropopause,
\item Constant $V_{CAS}$ above tropopause,
\item Constant $V_{CAS}$ below tropopause,
\item Acceleration in climb,
\item Deceleration in climb
\end{enumerate}
where $V_M$ is the Mach speed and $V_{CAS}$ is the calibrated speed. These flight conditions can be defined in terms of conjunctions of $q_1$, $q_2$ and $q_3$. From \cite{Bada2012} p.15-16, $f$ is discontinuous when $q$ jumps from one mode to another. Also, $f$ is bounded and therefore, with the flight envelope, one can also bound $\dot h$ and $\dot V$. One must pay attention to the Zeno behavior (cf. \cite{Lygeros2010}) where a system can make an infinite number of jumps in a finite amount of time. With these considerations in minds, one can use a numerical procedure to integrate eq.\ref{eq:roc} and eq.\ref{eq:acceleration}.

\subsection{Trajectory Generation}
In order to generate the trajectory, one must specify eq.\ref{eq:roc} and eq.\ref{eq:acceleration} as functions of time. With respect to BADA, let $Thr(h(t),t) = Thr(h(t))$, $D(h(t), V(t),t)=D(h(t),V(t))$ and $T(h(t),t) = T(h(t))$, that is the evolution of the thrust, the temperature and the drag are time-invariant. Moreover, the aircraft dynamic functions shall be specified with respect to the flight envelope constraints. 
In this study, we choose a nominal thrust function. In eq.\ref{eq:roc}, $Thr$ is replaced by the maximum thrust $Thr_{max}$ and the whole equation is multiplied by a reduced climb power coefficient $C_{red}$, which is supposed to give realistic profiles (cf. \cite{Bada2012} p.24).
To simulate the system, a common fourth-order Runge-Kutta method is used. Let $f_1(t,h,V,q) = \dot h(V,q)$ and $f_2(t,V,h, \dot h) = \dot V(h,\dot h)$. Then, one obtains the following integration scheme:
\begin{eqnarray}
  dh_1 &=& f_1(t_n,h_n,V_n,q(V_n,h_n)) \nonumber \\
  dv_1 &=& f_2(t_n,V_n,h_n, \dot h_n) \nonumber \\
  dh_2 &=& f_1( t_n + \frac{\Delta t}{2}, h_n + dh_1\frac{\Delta t}{2}, V_n + dv_1\frac{\Delta t}{2}, \nonumber\\
  && \quad q(V_n + dv_1\frac{\Delta t}{2},h_n + dh_1\frac{\Delta t}{2})) \nonumber\\
  dv_2 &=& f_2(t_n + \frac{\Delta t}{2}, V_n + dv_1\frac{\Delta t}{2},h_n + dh_1\frac{\Delta t}{2}, dh_2) \nonumber \\
  dh_3 &=& f_1(t_n + \frac{\Delta t}{2}, h_n + dh_2\frac{\Delta t}{2}, V_n + dv_2\frac{\Delta t}{2}, \nonumber\\
  && \quad q(V_n + dv_2\frac{\Delta t}{2},h_n + dh_2\frac{\Delta t}{2})) \nonumber\\
  dv_3 &=& f_2(t_n + \frac{\Delta t}{2},V_n + dv_2\frac{\Delta t}{2},h_n + dh_2\frac{\Delta t}{2}, dh_3) \nonumber \\
  dh_4 &=& f_1(t_n + \Delta t, h_n + dh_3 \Delta t, V_n + dv_3 \Delta t, \nonumber\\
  && \quad q(V_n + dv_3 \Delta t,h_n + dh_3 \Delta t)) \nonumber\\
  dv_4 &=& f_2(t_n + \Delta t, V_n + dv_3 \Delta t,h_n + dh_3 \Delta t, dh_4) \nonumber \\  
  h_{n+1} &=& h_n + \frac{\Delta t}{6} \left( dh_1 + 2dh_2 + 2dh_3 + dh_4 \right) \nonumber \\  
  V_{n+1} &=& V_n + \frac{\Delta t}{6} \left( dv_1 + 2dv_2 + 2dv_3 + dv_4 \right) \nonumber \\
  t_{n+1} &=& t_n + \Delta t \nonumber
\end{eqnarray}
where the initial conditions $t_0$, $q_0$, $h(t_0,q_0)=h_0 $, $V(t_0,h_0)=V_0$ are given. The choice of the Runge-kutta method is justified by the jumps in the function $f$. As a matter of fact, this method will minimize the impact of a jump during a timestep by averaging the variations at the beginning, twice at the middle point and at the end of the timestep (cf. \cite{Fortin2011}).
Nevertheless, stability questions shall be addressed in the later in order to validate the approach.

\subsection{Parameter Tuning}
Now that the trajectory generator is defined, we would like to tune the model parameters according to observations. First, there should be a trade-off between the number of parameters and the capacity of the model to approximate real trajectories. As a bad example, instead of using the default thrust controller defined in \cite{Bada2012}, we tried to find the optimal controller for the acceleration mode for a given trajectory. The resulting controller was a Bang-Bang control function where we have observed that the mode switches between accelerating and decelerating at every timestep. Clearly, this function can approximate any real trajectories, but does not generalize from one trajectory to another and does not reflect the real behavior of the aircraft. So, from eq.\ref{eq:roc} and eq.\ref{eq:acceleration} and the speed schedule $V^*$, the parameters $m$, $V_1$, $V_2$ ,$M$ and $\Delta T$ are good candidates since they are time-invariant, contrary to a discrete control law. Moreover, their values are fixed to nominal values and BADA explicitly suggests to tune them.

One way to tune the parameters is to minimize the position errors between the TP and the real trajectory. This leads to an optimization problem where the function to minimize is expressed by the trajectory generation scheme. Here, we decide to use a black-box optimization algorithm that will perform parameter estimation on the hybrid system. Due to the relations between the parameters and the complexity of the BADA models, we assume that parameter estimation is a non-trivial optimization problem.

\subsection{Black-Box Optimization}
As already mentioned, parameter tuning pertains to non-convex black-box optimization, and several methods could be used
to tackle it. Furthermore, no information whatsoever is available regarding the modality of the objective function with five parameters. As a matter of fact, it is unimodal when the mass parameter or the differential temperature are the only ones to be tuned. These have an effect on the whole trajectory. But, the speed parameters have a local effect depending on the speed schedule. Moreover, the differences between the speed parameters will induce acceleration phases that will transform the trajectory. Finding the best value for a speed parameter to fit locally the trajectory will create a local optimum that could be worse than finding the two speed parameters that will avoid an acceleration phase that is not undertaken in reality. Even if we did not prove that the objective function is multimodal, we can think that it can possibly be the case. Finally, the
objective is non-differentiable (or at least the analytical derivative
is out of reach). Hence general-purpose derivative-free optimization
method is required.

The Covariance Matrix Adaptation Evolution Strategy (CMA-ES)
\cite{hansenTutorial2005} is today a state-of-the-art derivative-free
optimization method that has demonstrated outstanding performances for
problems up to a few hundred variables, in several official
comparisons (see, among others, the CEC 2005 challenge
\cite{continuousCEC05}, and both Black Box Optimization Benchmark
workshops at ACM-GECCO 2009 \cite{BBOB09} and 2010 \cite{BBOB10}), as
well as on a large number of real-world applications
\cite{CMA-ESApplications}. CMA-ES is an Evolution Strategy
\cite{Rechenberg,Schwefel} that uses Gaussian mutation with adaptive
parameter setting. A Gaussian mutation is defined here by its
step-size and its covariance matrix. The step-size is increased
(resp. decreased) if the cumulated path of the current best solution
is smaller (resp. larger) than that of a random walk, and in the
original version \cite{Hansen:ICEC96,Hansen:ECJ01}, the covariance
matrix was updated by adding a rank-one matrix with eigenvector the
direction of progress. An improved version with rank-$\mu$ update was
later proposed \cite{Hansen:ECJ03}, and several additional variant
made it more and more powerful. The most recent version is the
so-called bi-pop-CMA-ES \cite{bipopCMA2009}, that evolves both a large
and a small population and outperforms previous versions in case of
multi-modal functions. All source code is available on the author's
web page (\url{http://www.lri.fr/~hansen/index.html}), in different
programming languages, including the bi-pop version. Using CMA-ES for parameter estimation is then straightforward, and amounts to
interfacing the objective function for BADA TP with the core CMA-ES
program, after eventually normalizing its parameters.

\section{Model Validation}
\label{sec:modelValidation}
This section deals both with the empirical validation of the nominal model and the estimation algorithm used to fit the parameters on the whole trajectory.

\subsection{Dataset}
\label{subsec:dataset}
To validate our model, we will use a dataset composed of 262 real departure trajectories of A320. These trajectories have been recorded via a radar systems during one month. For one trajectory, there is a data vector at every 5 seconds composed of the aircraft position, the rate of climb and the true airspeed. Then, the top of climb is calculated as the first highest point of the trajectory. Also, these trajectories do not have a rate of climb that is equal to zero for more than 30 seconds, they shall reach a cruise level at least of 300 FL and their durations are at least of 1100 seconds. These filtering conditions permits to keep trajectories that are not affected by a level flight clearance.

\subsection{Methodology}
\label{subsec:method}
To assess the model, we evaluate the position error between the TP and the real trajectory. Let $\mathcal{H}(\theta;s_0)$ be the sequence of altitudes generated by the simulation of the hybrid system with the parameters $\theta = \left[ m \quad \Delta T \quad V_1 \quad V_2 \quad V_M \right]^T$ and the initial condition vector $s_0 = \left[ t_0 \quad q_0 \quad h_0 \quad V_0 \right]^T$. Then, the measure is simply the sum of absolute errors. In the case of parameter estimation, we search in the feasible space of parameters $\Theta$, the point $\theta \in \Theta$ that minimizes this measure. So, we have the optimization problem: 
\begin{equation}
\label{eq:opt}
\theta_{i,j}^* = \underset{\theta \in \Theta}{\operatorname{argmin}} \sum_{n=i}^j \left| \mathcal{H}(\theta;s_0)_n - T_n \right|
\end{equation}
where $\mathcal{H}(\theta,s_0)$ is the sequence of ordered altitudes obtained for given parameters $\theta$ and initial conditions $s_0$ and $T$ is the sequence of observed altitudes from a trajectory of the dataset. We suppose that $(j-i) \leq |\mathcal{H}(\theta,s_0)| \leq |T|$ and that the timestamp associated to the point $\mathcal{H}(\theta,s_0)_n$ is the same than $T_n$. When fitting the whole climbing phase, some difficulties with eq.\ref{eq:opt} may arise. First, we need to give a termination criterion when simulating the hybrid system that is to stop when the TP reaches the level flight. Depending on the parameters, the cardinal of the resulting sequence of points will be different. So, it might be necessary to add points to the real trajectory T since the TP can reach the top of climb after T, e.g. a high value for the mass parameter. To compute this error, we simply add points at the level flight until the TP reaches it. Inversely, we add these points to the TP if it reaches the top of climb before the real trajectory. 
Besides, $i < j$ are bounds to restrain the optimization on any subset of contiguous points of the trajectory. Thereafter, this notation will be useful for the online predictor. 

\begin{table}
\centering
\renewcommand{\arraystretch}{1.3}
\caption{\label{tab:badaVad} Modelization Errors - Mean and Standard Deviation}
\begin{tabular}{|c|c|c|}
  \hline
  Time after takeoff & Nominal (FL) & Tuned (FL) \\ \hline 
  2min. & 4.9195 (3.1422) & 3.0929 (2.3133) \\ \hline
  5min. & 7.1416 (4.8556) & 2.5496 (2.5282) \\ \hline
  10min. & 9.6714 (6.6146) & 1.4057 (1.7441) \\ \hline
  15min. & 10.9441 (9.0016) & 2.1957 (2.2600) \\ \hline
  20min. & 11.8008 (8.8068) & 2.0546 (2.1367) \\ \hline
\end{tabular}
\end{table}

\subsection{Results}
\label{subsec:result}
\begin{figure}[!t]
\centering
\includegraphics[width=2.5in]{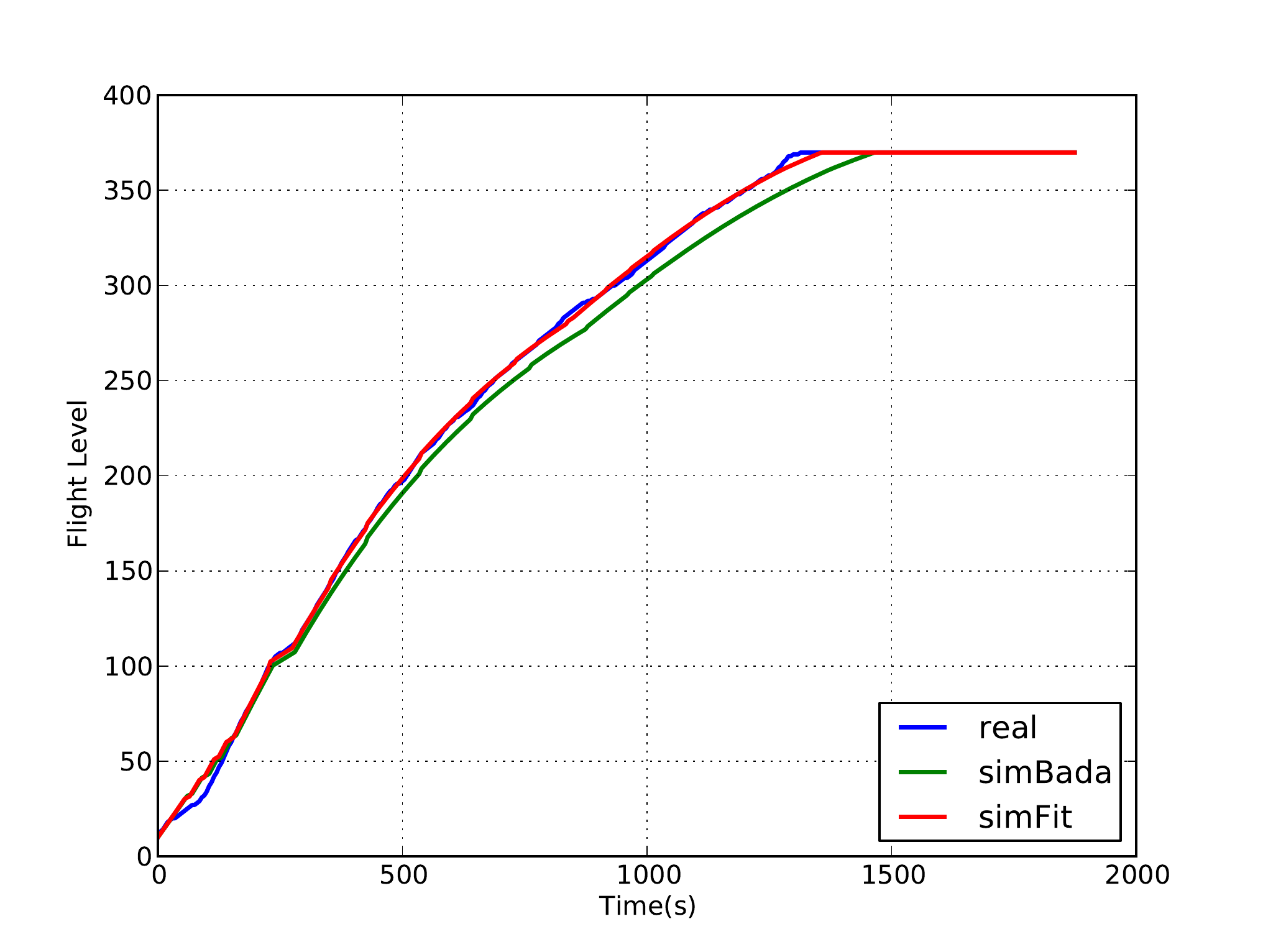}
\caption{Trajectory Fitting}
\label{fig:traj}
\end{figure}

\begin{figure}[!t]
\centering
\includegraphics[width=2.5in]{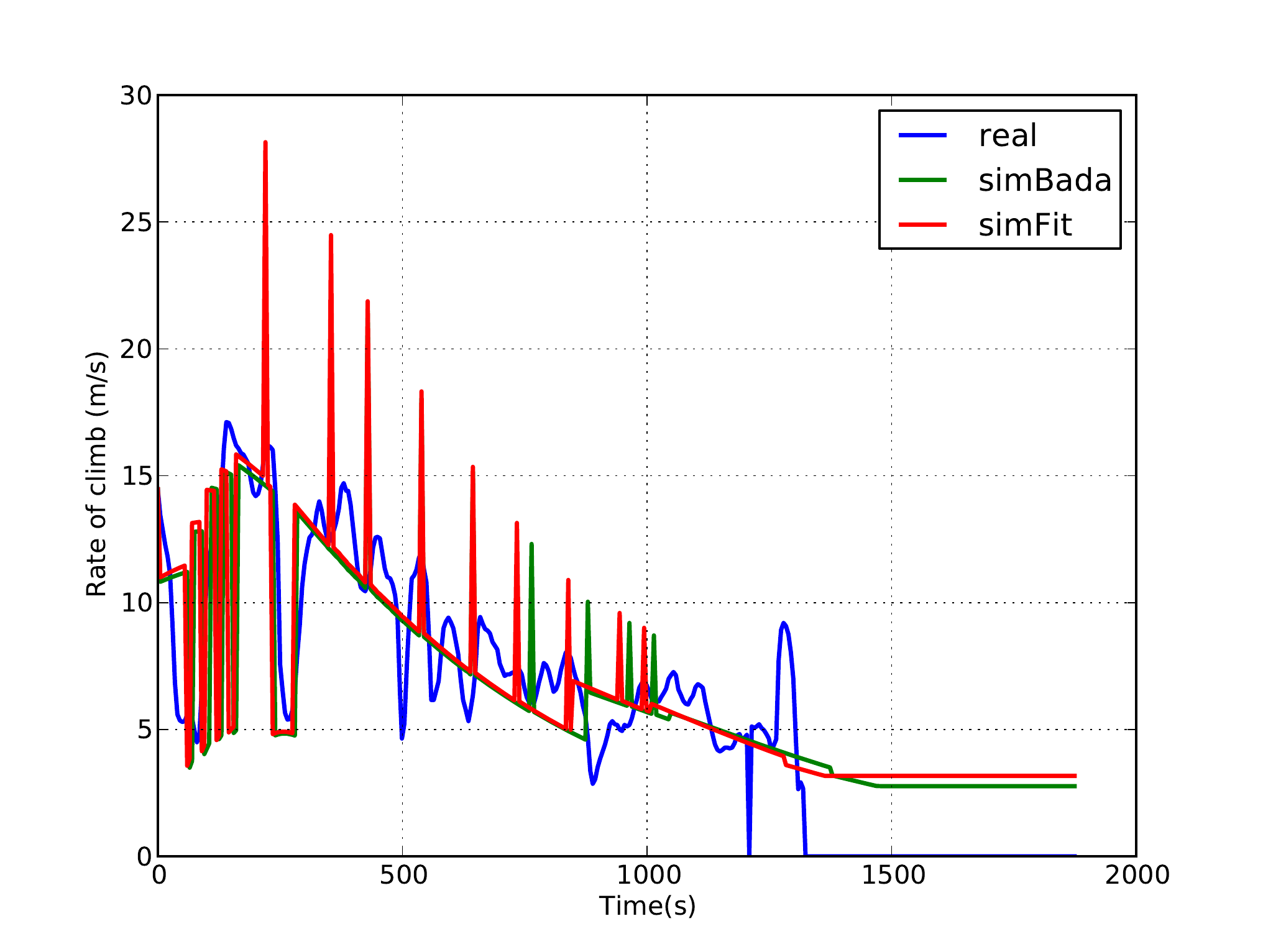}
\caption{Rate of Climb Modeling}
\label{fig:roc}
\end{figure}

Table \ref{tab:badaVad} shows the evolution of the mean errors and the standard deviation with time for both models: with nominal values and with tuned values. The first important evidence is the inaccuracy of the Total-Energy Model to model the positions at the beginning of the trajectory. As a matter of fact, for the tuned values, the errors at 2 minutes are the highest. 
An explanation consists in the fact that the aircraft states (position, speed, heading) change rapidly at the beginning of the trajectory and the selected five parameters are not sufficient to capture this complexity.
Furthermore, the optimization of the equation eq.\ref{eq:opt} has a global scope and so, the errors generated by local behaviors of the aircraft are ignored in favor of the common behavior.
This depends directly on the selected values of $i$,$j$ in $\theta_{i,j}^*$, which in this case are $i=0$ and $j= t_{toc}$ where $t_{toc}$ is the time at top of climb.
From our dataset, this common behavior happens around 10 minutes where the errors are the smallest after an acceleration phase which happens around 5 minutes.
As an example, on the figure \ref{fig:traj}, we can distinguish three main behaviors: the initial climb from 0s. to approximately 200s, a short acceleration phase from 200s. to 300s. and the common behavior after 300s.
From figure \ref{fig:roc}, we can see that the initial climb is characterized with high variability in the rate of climb and an acceleration phase between 50s. and 100s., shown by a huge decrease in the rate of climb.
During this phase, the predicted rate of climb is far from reality for both parameter sets.
Thereafter, both models capture the acceleration phase and finally, they average the rate of climb during the common behavior, which fits well the positions as shown on figure \ref{fig:traj}.

Another interesting result is the evolution of the standard deviation, which is in parentheses, for the model with nominal values. 
It increases with time from 2 min. to 15 min. and afterwards, it seems to stabilize around 9 FL. 
This can be interpreted as an uncertainty cone, which is often used in the air traffic community, but here, we can see that the cone stops to grow at 15 minutes and becomes a corridor of uncertainty.
First, we understand that the cone grows with the flight envelope, but afterwards, the uncertainty is bounded by the fact that the trajectories, as functions of time, are strictly increasing because of the filtering conditions of the dataset (cf. Section \ref{subsec:dataset}) and the upper bound that is the cruise level.

\section{Online Trajectory Predictor}
\label{sec:Online}
In this section, we present an online trajectory prediction which uses the observed positions of a current flight to tune at the same time the parameters of the model used to predict the rest of the trajectory.
Different kind of algorithms can be used to undertake this task. 
Traditionally, a probabilistic approach is used in this online configuration, e.g. Kalman Filter, where the uncertainty is explicitly modeled. 
These performs very well on short periods of time but, for longer period like in this context, the linearity of the model could be too limited and should be subject to future experiments.
In our approach, we use the BADA model in conjunction with the optimization algorithm CMA-ES in order to fit the parameters of the model with an objective function that defines a distance between the observed positions and the modeled one. As we will see, the problem of overfitting, which is studied by statistics and machine learning, arises in our context.
The current section presents the difficulties and the solutions chosen with the associated results.

\subsection{Design of Experiment}
In order to verify that the idea of an online predictor, as the one mentioned previously, is valid, we must do an empirical evaluation of the chosen algorithm by replicating the same context.
The main hypothesis is that from the observation set, we can determine the values of parameters that will be fitted to the current flight. 
So the trajectory is separated in two subsets: the observed altitudes from the beginning to the present and the future altitudes from the present to the top of climb.
As in the subsection \ref{subsec:result}, we use metering points to evaluate the quality of the prediction by computing the errors between the predicted altitudes and the real ones.
To distinguish if a set of errors is statistically greater than the other one, we use a Wilcoxon signed-rank test. 
The null hypothesis of this test is that two related paired samples come from the same distribution. 
In our case, this test is a relevant choice because two approaches are tested on the same trajectory dataset. 
As usual, we reject the null hypothesis if the p-value is lower than 0.05.

\subsection{Methodology}
The most naive way to learn the parameters of the model from the observed altitudes is to directly apply Eq.\ref{eq:opt} from subsection \ref{subsec:method} and to apply them to generate the rest of the trajectory.
By doing so, the default model is always better than the fitted one with a significant p-value. 
The reason behind this result is simply that the fitted model does not generalize over all the behaviors of the aircraft. 
In other words, it is fitted only to the behavior captured in the observations. This problem is referred to overfitting in Machine Learning litterature.
To circumvent this problem, we must think of the trajectory as a time serie where the observations arrive with an determined order, i.e. the temporal order.
So, at first, we will always observe the initial climb where we know from Table \ref{tab:badaVad} that the inaccuracy of the model is the greatest.
Furthermore, we are more interested by the parameter values that fit better the positions near the present time than at the beginning of the trajectory.
To this end, we will add a weighting vector $\alpha$ that will penalize more the errors that are near the present time.
But still, this is not sufficient because, depending on the present time, some parameters will not have any effect on the trajectory.
As a matter of fact, from 0 to FL60, a predefined schedule is applied and only the mass parameter has an effect in BADA. 
The scope of the parameter $V_1$ is from FL60 to FL100, the scope of $V_2$ is from FL100 to the transition altitude (around FL277) and finally, the scope of $V_m$ is over the transition altitude. 
Furthermore, we add the constraint that $V_2$ is greater than $V_1$ to the optimization problem. 
To avoid that the optimization algorithm assigns them some arbitrary values resulting in unrealistic trajectories, we use a regulation method that penalizes any deviation from the default parameters.
A meta-parameter $\lambda$ is associated to the weight of the penalty, which controls the tradeoff between exploration and exploitation.
Consequently, the resulting objective function is:

\begin{equation}
\label{eq:learning}
\theta_{0,t}^* = \underset{\theta \in \Theta}{\operatorname{argmin}} \left[ \sum_{i=0}^t \alpha_i \left| \mathcal{H}(\theta;s_0)_i - O_i \right| + \lambda \sum_{i=0}^{|\theta|} \left| \theta_i - \theta_i^d \right| \right] 
\end{equation}

where $\theta^d$ is the default parameters of BADA. Notice that the penalty occurs on the parameters space, but could also be applied on the trajectory space because the mapping from the parameter space to the trajectory space is not linear in the sum of differences of altitudes. 

\begin{figure}[!t]
\centering
\includegraphics[width=2.5in]{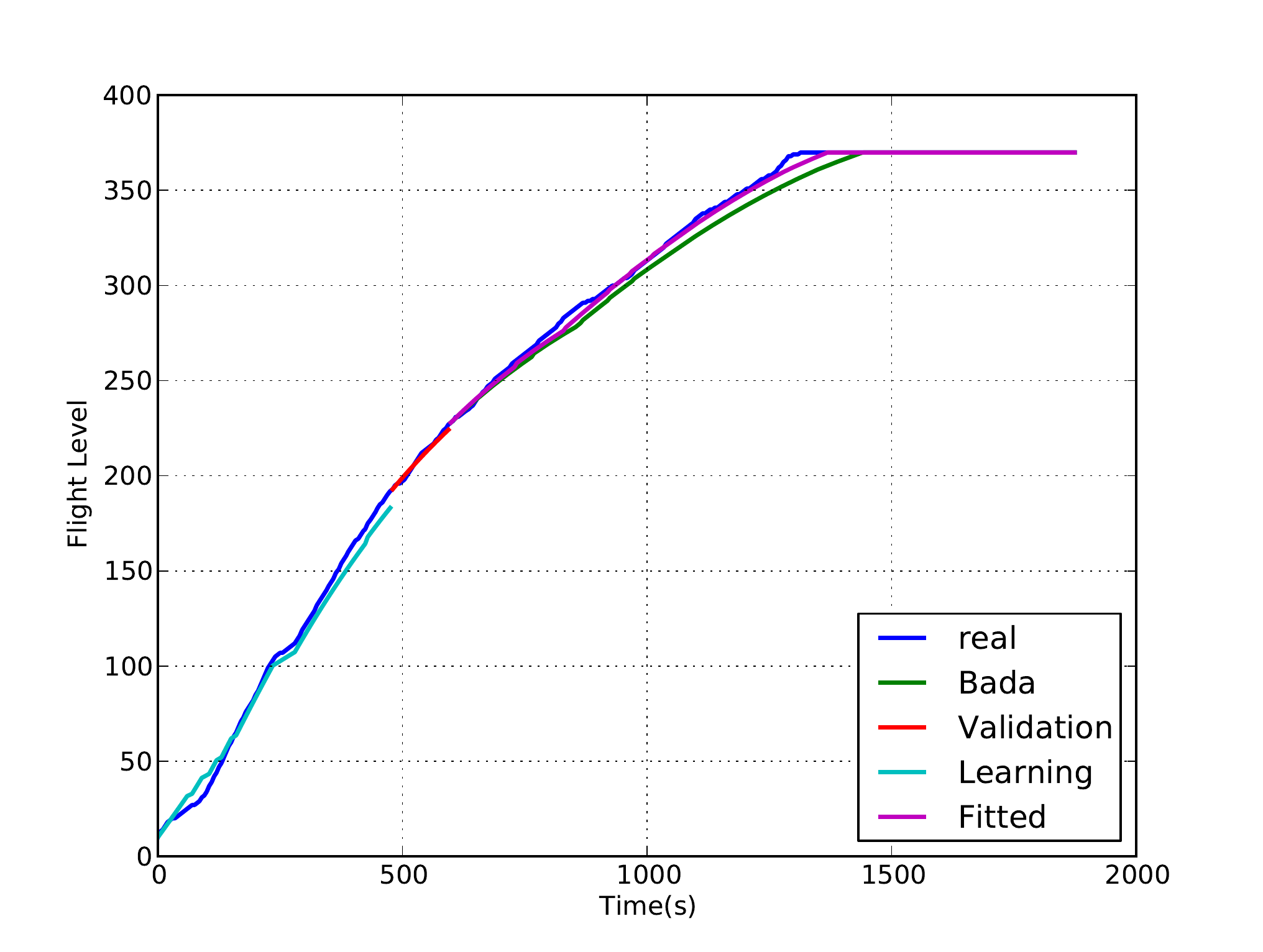}
\caption{Online Prediction}
\label{fig:online}
\end{figure}

Finally, in order to set the value of $\lambda$, we use a cross-validation approach where we partition the observation set in two: the learning set and the validation set. 
We choose the validation set to be just before the current altitude. 
One must notice that the samples are not independent and identically distributed and that we create a bias in favor of the points located just after the current altitude.
Because of our extrapolation context, a bias is inevitable and this one seems the most justifiable one in order to gain accuracy in predicting the future positions.
Figure \ref{fig:online} shows the partition of the trajectory.
The cross-validation technique used in this study consists in learning the parameters of the model on the learning set with the objective function and to use these parameters for generating the points on the validation intervals. 
Then, we compare these points with the real ones. 
We do it for multiple values of $\lambda$ and we choose the parameter values where the validation error is the lowest to generate the rest of the trajectory.

\subsection{Results}
The approach is validated on the same dataset than subsection \ref{subsec:dataset}.
We choose three different time slices in order to represent the online aspect of the method.
The validation set size is fixed to 36 points or 180 seconds.
This choice must do the trade-off between the validation purpose of avoiding overfitting which could lead to a large validation set size and the learning purpose of finding the best parameter values which could lead to a large learning set size and therefore, a small validation set size. 
At least, the validation set size must be higher than the acceleration phase, where the local behavior is the most different from the global one.
In the learning objective function (cf. Eq.\ref{eq:learning}), we choose a linear weight function where $\alpha_i = \frac{i}{t-1}$.
The initial lambda value is arbitrarily set to 100 and are doubled until the penalty is high enough so that the fitted values equal the default ones.
Then, the parameter values generating the lowest validation error are chosen.

Also, to avoid that the algorithm changes the parameter values based on a poor learning performance, we arbritrarily set a threshold error at 5 FL, which is higher than the results at subsection \ref{subsec:result}. 
When the threshold is exceeded, the BADA default values are chosen.

Table \ref{tab:onlineVadP400}, \ref{tab:onlineVadP500} and \ref{tab:onlineVadP600} show the results of the proposed methodology for each time slice.
At $P=400$s, the two model performances are not significantly different as shown with the high p-values. 
We can see that our approach increases the accuracy by 1 FL in average for the metering points at 2 minutes and 5 minutes after the current time slice with a p-value significantly under 0.05. 
For the metering points at 10 minutes, the p-value is higher than 0.05 and the difference is not statistically significant due to the high standard deviation values.
This can be explained by the fact that the model is not very accurate during initial climb (cf. model validation section) and the validation set covers the acceleration phase. 
Also, because the learning error is too high, the algorithm can choose the BADA default values.
So, the default choice ratio is around $20\%$ which is rather high.

At $P=500$s, the fitted model performs better at 2 and 5 minutes after the current position with small p-values.
For 10 minutes, the two models are not significantly different because of the high value of the standard deviation. 
In fact, this can be interpreted as the two models are equally affected by the uncertainty around the possible maneuvers of the aircraft.
In order to perform better, more information on the flight intents are required to reduce the variability in the trajectories.
Here, the ratio of the default choice is $16\%$.

At $P=600$s, the results are similar to $P=500$s.  
The reason is that the aircraft keeps the same behavior between 500s. and 600s. which is different from the behavior at 400s.
There is some kind of regularity that explains the fact that the prediction is enhanced up to 5 minutes. 
This regularity is captured more easily by the learning algorithm when the behavior is stable during the validation interval.
In this case, the ratio of the default choice is $14\%$.

\begin{table}[h!]
\centering
\renewcommand{\arraystretch}{1.3}
\caption{\label{tab:onlineVadP400} Comparison of Online Models at $P=400$s}
\begin{tabular}{|c|c|c|c|}
  \hline
  Time after takeoff & Nominal (FL) & Tuned (FL) & p-value \\ \hline 
  2min. & 3.3029 (2.6698) & 3.1699 (2.6740) & 0.3401\\ \hline
  5min. & 6.7553 (5.6084) & 6.5518 (5.6578) & 0.6726\\ \hline
  10min. & 8.7851 (7.0757) & 9.1846 (7.5687) & 0.4541\\ \hline
\end{tabular}
\end{table}

\begin{table}[h!]
\centering
\renewcommand{\arraystretch}{1.3}
\caption{\label{tab:onlineVadP500} Comparison of Online Models at $P=500$s}
\begin{tabular}{|c|c|c|c|}
  \hline
  Time after takeoff & Nominal (FL) & Tuned (FL) & p-value\\ \hline 
  2min. & 4.0406 (3.2758) & 3.2834 (2.7237) & 5.612e-4\\ \hline
  5min. & 8.1290 (6.0885) & 7.0567 (4.8281) & 0.02049\\ \hline
  10min. & 9.0872 (7.0085) & 9.4205 (6.7658) & 0.7939\\ \hline
\end{tabular}
\end{table}

\begin{table}[h!]
\centering
\renewcommand{\arraystretch}{1.3}
\caption{\label{tab:onlineVadP600} Comparison of Online Models at $P=600$s}
\begin{tabular}{|c|c|c|c|}
  \hline
  Time after takeoff & Nominal (FL) & Tuned (FL) & p-value\\ \hline 
  2min. & 4.5110 (3.4354) & 3.5912 (2.4845) & 1.289e-05\\ \hline
  5min. & 6.7936 (4.9209) & 5.7231 (4.0456) & 1.289e-03\\ \hline
  10min. & 8.5131 (6.6410) & 9.4992 (7.8805) & 0.09098\\ \hline
\end{tabular}
\end{table}

\section{Conclusions and Future Work}
\label{conclusion}
In conclusion, this article presents a flight model for the climbing phase defined as a hybrid system based on BADA. 
An integration scheme is defined in order to generate the trajectories from this system. 
Then, tuning parameters are identified in order to be used in the online context. 
The method is validated through measuring the vertical errors between real trajectories and generated ones: both for default and fitted parameter values. 
In the validation context, fitting is done on the entire trajectory i.e. with total information. 
The measured errors are considered as the accuracy limit of this five parameters model. 
Afterwards, the model is applied in the online context, which evolves with time. 
Known altitudes are used to fit the parameters and then, these are used to predict the remaining points. 
To avoid overfitting, the known points are partitioned  in a learning set and a validation set. 
The validation set is used to fit the regulation parameter, which penalizes the deviation from the default values. Results shows that the initial climb, which is before the main acceleration phase, is not modeled accurately in order to fit the parameters. 
On the contrary, when the flights adopt a common behavior after this acceleration phase, the online learning method increases the accuracy of the trajectory prediction. 
The gain is about 1 FL for 2 minutes and 5 minutes after the current time. 
After that, the two models are not significantly different because of the huge uncertainty on the trajectory. 
This study shows that the uncertainty becomes too important between 5 and 10 minutes with minimal information. 
Consequently, without further information on the flight intents and airspace constraints, ground trajectory prediction is not accurate enough for automated tasks such as conflict resolution.
Furthermore, this study confirms the need to use the aircraft derived data to feed the BADA model in order to build a ground trajectory prediction as the foundation stone of automated Air Traffic control systems. 
The next step should be the identification of the relevant onboard data improving significantly the trajectory predictability on ground. 

\section{Acknowledgments}
This author Ga{\'e}tan Marceau is funded by the scholarship CIFRE 710/2012 established between Thales Air Systems and INRIA-Saclay, and the scholarship 141138/2011 from the {\it Fonds de Recherche du Qu{\'e}bec - Nature et Technologies}.

\bibliographystyle{plain}
                  \bibliography{library2,Evolution}

\end{document}